\pdfoutput=1

\documentclass[11pt]{article}
\usepackage[dvipsnames]{xcolor}

\usepackage[final]{acl}

\usepackage{times}
\usepackage{latexsym}

\usepackage[T1]{fontenc}

\usepackage[utf8]{inputenc}

\usepackage{microtype}

\usepackage{inconsolata}

\usepackage{graphicx}
\usepackage{amsmath}

\usepackage[utf8]{inputenc} 
\usepackage[T1]{fontenc}    
\usepackage{hyperref}       
\usepackage{url}            
\usepackage{booktabs}       
\usepackage{amsfonts}       
\usepackage{nicefrac}       
\usepackage{microtype}      
\usepackage{todonotes}
\usepackage{listings}
\usepackage{tabularx}
\usepackage[normalem]{ulem}
\usepackage{multirow}
\usepackage[export]{adjustbox}

\newcolumntype{C}[1]{>{\centering\arraybackslash}p{#1}}
\newcolumntype{L}[1]{>{\raggedright\arraybackslash}p{#1}}

\title{FactSelfCheck: Fact-Level Black-Box Hallucination Detection for LLMs}

\author{Albert Sawczyn\textsuperscript{1} \And  Jakub Binkowski\textsuperscript{1} \And Denis Janiak\textsuperscript{1} \AND Bogdan Gabrys\textsuperscript{2} \And Tomasz Kajdanowicz\textsuperscript{1} \AND
        \vspace{-1.5em} \\ \textsuperscript{1}Wrocław University of Science and Technology \\ 
        \textsuperscript{2}University of Technology Sydney \\ \vspace{-1em} \\ albert.sawczyn@pwr.edu.pl}

\def\methodname{FactSelfCheck}

\begin{document}
\maketitle

\begin{abstract}
Large Language Models (LLMs) frequently generate hallucinated content, posing significant challenges for applications where factuality is crucial. While existing hallucination detection methods typically operate at the sentence level or passage level, we propose \nobreak{FactSelfCheck}, a novel zero-resource black-box sampling-based method that enables fine-grained fact-level detection. Our approach represents long-form text as interpretable knowledge graphs consisting of facts in the form of triples, providing clearer insights into content factuality than traditional approaches. Through analyzing factual consistency across multiple LLM responses, we compute fine-grained hallucination scores without requiring external resources or training data. Our evaluation demonstrates that \nobreak{FactSelfCheck} performs competitively with leading sentence-level sampling-based methods while providing more detailed and interpretable insights. Most notably, our fact-level approach significantly improves hallucination correction, achieving a $35.5\%$ increase in factual content compared to the baseline, while sentence-level SelfCheckGPT yields only a $10.6\%$ improvement. The granular nature of our detection enables more precise identification and correction of hallucinated content. Additionally, we contribute FavaMultiSamples, a novel dataset that addresses a gap in the field by providing the research community with a second dataset for evaluating sampling-based methods. 
\end{abstract}

\section{Introduction}
\label{sec:introduction}

Large Language Models (LLMs) have gained significant attention from academia and industry recently. However, a major limitation of LLMs is their tendency to generate hallucinated information \citep{farquhar_detecting_2024,10.1145/3703155_survey}, posing significant challenges for applications where factual correctness is crucial, such as healthcare \citep{healthcare11060887}. Although numerous methods have been proposed to reduce hallucinations \citep{zhang2023sirenssongaiocean}, it is not possible to eliminate them, and LLMs will constantly hallucinate \citep{math11102320, xu2024hallucinationinevitableinnatelimitation}. Therefore, there remains a critical need for reliable hallucination detection in LLM responses, particularly for long-form text generation tasks where complexity and information density increase the risk of factual errors. Effective detection enables system interventions by either preventing the transmission of hallucinated content to users or facilitating its correction \citep{zhang2023sirenssongaiocean}.

Previous approaches to hallucination detection have primarily focused on classifying hallucinations at either the passage or sentence level \citep{10.1145/3703155_survey}. While valuable, these approaches are limited in their granularity and interpretability, as they do not provide detailed information about specific hallucinated facts or clear insights into what exactly is wrong. This limitation becomes particularly evident in long-form text generation. To address this limitation, we propose a novel method for hallucination detection that operates at the fact level, offering finer-grained and more interpretable analysis. In our approach, we define a fact as a triple consisting of a head, relation, and \mbox{tail -- a standard} representation in knowledge graphs (e.g., (\textit{Robert Smith},~\textit{member of},~\textit{The Cure})) \citep{hamiltonRepresentationLearningGraphs2017}. Our method provides more precise, actionable, and interpretable information by computing hallucination scores for individual facts than traditional passage-level or sentence-level classification approaches. This structured representation, through knowledge graphs, provides enables straightforward interpretation and verification \citep{pan-etal-2024-automatically}.

\begin{figure*}[!htb]
    \centering
    \includegraphics[trim=0cm 0.3cm 1cm 0.5cm,clip=true,width=0.9\textwidth]{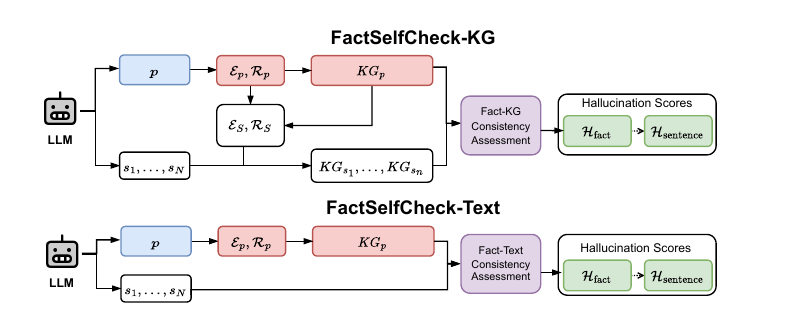}
    \caption{The pipeline of \methodname~in two variants. For response $p$, entities $\mathcal{E}_p$ and relations $\mathcal{R}_p$ are extracted, followed by the construction of knowledge graphs $KG_p$, for which hallucination scores $\mathcal{H}_{\text{fact}}$ are calculated. Samples' entities $\mathcal{E}_S$ and relations $\mathcal{R}_S$ are created by merging $\mathcal{E}_p$ and $\mathcal{R}_p$ with entities and relations from $KG_p$. For each sample $s$, the knowledge graph $KG_s$ is extracted. FactSelfCheck-KG assesses the consistency between a fact and all $KG_s$. FactSelfCheck-Text assesses the consistency between a fact and all $s$ directly. To obtain sentence-level score $\mathcal{H}_{\text{sentence}}$, fact-level scores are aggregated, as indicated by \dashuline{dashed arrows}.}
    \label{fig:method}
    \end{figure*}

Our granular approach is motivated by two key observations. First, a single sentence can contain multiple facts, with the number of facts varying significantly across sentences, contexts, and domains. This variability makes it challenging to identify hallucinated aspects of generated output precisely when using sentence-level detection. Second, false information can be dispersed throughout a text, as a single fact may appear across multiple sentences. That can mislead the sentence-level detection, as the factuality of a sentence is dependent on the previous sentences \citep{10.5555/3692070.3694535snowball}. Fine-grained fact-level detection provides a more precise understanding of text factuality than a sentence analysis. It enables better assessment of content reliability and, as we show later, more effective factuality correction.

We propose FactSelfCheck, a black-box method for fact-level hallucination detection, meaning it does not require access to the model's internal parameters. This design choice makes our approach universally applicable across any LLM, including closed models, like GPT \citep{openai2024gpt4ocard}. Following a sampling-based detection paradigm, introduced by \citet{manakul-etal-2023-selfcheckgpt}, our method utilizes multiple response generations and analyzes the factual consistency of extracted facts across these samples. This paradigm is based on the phenomenon that factual information remains largely consistent across different generations, while hallucinated content tends to vary or contradict itself between samples \citep{manakul-etal-2023-selfcheckgpt, DBLP:conf/iclr/0002WSLCNCZ23_self_consistency}. This way, we can effectively identify hallucinated facts without relying on external resources (zero-resource) or access to the model's internal parameters (black-box). Our zero-resource, non-parametric approach requires no external knowledge bases or training data, making it broadly applicable across domains. The FactSelfCheck pipeline consists of three main steps: knowledge graph extraction, which extracts sets of facts from the initial response and samples; fact-level hallucination scoring; and calculating sentence-level scores by aggregating fact-level scores.

We evaluated our method using the WikiBio GPT-3 Hallucination Dataset \citep{manakul-etal-2023-selfcheckgpt} and the FavaMultiSamples dataset. WikiBio was the only existing dataset for evaluating sampling-based hallucination detection methods. To address this significant gap, we developed FavaMultiSamples, providing an additional evaluation benchmark. We performed both sentence-level and fact-level evaluation, with sentence-level scores aggregated from fact-level scores. Our approach achieves performance comparable to leading sampling-based methods at the sentence level while providing more detailed information about hallucinations. We also demonstrate effective fact-level hallucination detection and show that our fact-level approach significantly improves hallucination correction. Compared to a baseline, providing incorrect facts to the correction method leads to a $35.5\%$ increase in factual content, while passing incorrect sentences leads to an $10.6\%$ increase.

\paragraph{Our key contributions are as follows:}\label{contributions}
\begin{enumerate}
    \item The novel zero-resource black-box sampling-based method for fact-level hallucination detection -- FactSelfCheck, designed for long-form text generation. It enables fine-grained hallucination detection in LLM responses without requiring training data or external resources, as it is both non-parametric and zero-resource. We propose two effective approaches for measuring factual consistency across multiple samples: FactSelfCheck-KG using knowledge graph comparisons and FactSelfCheck-Text using direct text comparison.
    \item The FavaMultiSamples dataset, a novel dataset for evaluating sampling-based methods.
    \item Comprehensive evaluation of our method, which shows competitive performance with leading sampling-based methods while providing more detailed insights.
    \item Demonstration that fact-level detection significantly improves hallucination correction compared to sentence-level approaches.
\end{enumerate}

Our code is available on GitHub \footnote{\href{https://github.com/graphml-lab-pwr/FactSelfCheck}{github.com/graphml-lab-pwr/FactSelfCheck} license: \nobreak{CC~BY-SA~4.0}}. The \nobreak{FavaMultiSamples} dataset is available on Hugging Face \footnote{\href{https://huggingface.co/datasets/graphml-lab-pwr/FavaMultiSamples}{huggingface.co/datasets/graphml-lab-pwr/FavaMultiSamples} license: \nobreak{CC~BY-SA~4.0}}. We publish all the code and data, allowing for the reproduction of the results.

\section{Related work}
\label{sec:related}

\citet{xu2024hallucinationinevitableinnatelimitation} have proven that hallucinations are inevitable in LLMs. As LLMs are powerful tools, many recent studies have been conducted regarding hallucination mitigation and detection \citep{zhang2023sirenssongaiocean, 10.1145/3703155_survey}. The detection methods can be divided into two groups: white-box and black-box.

White-box methods analyze LLMs' internal states \citep{farquhar_detecting_2024, azaria_internal_2023}. While these methods are universal across all LLMs, they often require multiple generations, similar to sampling-based methods. Notable approaches include: SAPLMA \citep{azaria_internal_2023}, which predicts from hidden states whether generated text is correct or incorrect; INSIDE \citep{chen2024insidellmsinternalstates}, which evaluates hidden state consistency across generations; SEPs \citep{kossen2024semanticentropyprobesrobust} that predict entropy directly from model hidden states; Lookback Lens \citep{chuang-etal-2024-lookback} and AttentionScore \citep{sriramanan2024llmcheck} that uses attention maps to detect hallucinations.

Black-box approaches operate without access to the model's internal states and aim to detect hallucinations based solely on the text generated by LLMs. Some of these methods use external resources to collect evidence \citep{min-etal-2023-factscore, chern2023factoolfactualitydetectiongenerative}. Others leverage LLMs to detect hallucinations like CoVe \citep{dhuliawala-etal-2024-chain}, which utilizes the chain-of-thought paradigm for detection. Another category is sampling-based methods, such as SelfCheckGPT \citep{manakul-etal-2023-selfcheckgpt}, which evaluate factuality by generating multiple responses (stochastic samples) and assessing consistency between the original response and these samples. The paradigm of utilizing LLM to check its own responses was widely studied and adopted in many works \citep{kadavath2022languagemodelsmostlyknow,lin-etal-2024-criticbench, palmeira-ferraz-etal-2024-llm, zhang-etal-2024-self, miao2023selfcheckusingllmszeroshot}. Many of these approaches employ a multi-step decomposition strategy to break down the complex task of hallucination detection into more manageable subtasks, a methodology we also adopt in our approach.

The most popular approach is to classify hallucinations at sentence-level or passage-level \citep{10.1145/3703155_survey}. Few methods have been specifically designed to detect hallucinations at the fact level, i.e. where facts are defined as triples. GraphEval \citep{sansford2024graphevalknowledgegraphbasedllm} generates a KG from LLM output and compares it with the context provided in the LLM input. FactAlign \citep{rashad-etal-2024-factalign} builds KGs from LLM output and source text, then compares them after performing entity alignment, a technique that pairs the same entity in different KGs. Knowledge-centric detection \citep{hu-etal-2024-knowledge} similarly extracts knowledge triplets for fine-grained detection. All these methods require external context as reference, while our method is designed to work without any external knowledge sources by leveraging sampling-based consistency analysis. When evaluating on well-established sources like Wikipedia, methods with access to source materials may achieve superior performance \citep{manakul-etal-2023-selfcheckgpt}. However, our zero-resource approach offers broader applicability, as it can be applied to any task without requiring external knowledge sources. 

Most similar to our approach is GCA \citep{fang2024zero}, which constructs KGs from the response and samples and then compares them by aggregating multiple scores. However, GCA has significant methodological concerns -- they tuned 6 hyperparameters directly on the evaluation set (the only available split in the WikiBio GPT-3 hallucination dataset \citep{manakul-etal-2023-selfcheckgpt}), making it methodologically problematic. Due to these concerns, we cannot provide a fair quantitative comparison with GCA. In contrast, FactSelfCheck is truly zero-shot and parameter-free, requiring no parameter tuning. We achieved that by designing a constrained KG extraction that works consistently across multiple generation samples, rather than using freeform extraction like GCA.

\section{Method}
\label{sec:method}

We propose FactSelfCheck, a black-box sampling-based method for fact-level hallucination detection, as illustrated in Figure \ref{fig:method}. Our method is specifically designed for long-form text generation scenarios, where passages typically contain several sentences with complex factual information.

\subsection{Notation}

Let $p$ denote the initial response passage generated by the LLM to a user query, which we aim to evaluate for hallucinations. Let $S = \{s_1, \ldots, s_N\}$ represent a set of $N$ stochastic LLM response samples. The text passage $p$ consists of a set of sentences $U$. For each sentence $u \in U$ and each sample $s \in S$, we extract knowledge graphs $KG_u$ and $KG_s$, respectively. Each knowledge graph comprises a set of facts, where a fact $f$ is defined as a triple $(h, r, t)$ consisting of a head $h$, relation $r$, and tail $t$, e.g. (\textit{Robert Smith}, \textit{member of}, \textit{The Cure}). We define $KG_p = \bigcup_{u \in U} KG_u$ as the knowledge graph consisting of all facts from the passage $p$.

Our objective is to compute a fact-level hallucination score $\mathcal{H}_{\text{fact}}$ for each fact $f$ in $KG_p$. Subsequently, to facilitate comparisons with other methods, we aggregate these scores to obtain a sentence-level hallucination score $\mathcal{H}_{\text{sentence}}$ for each sentence $u$.

\subsection{FactSelfCheck pipeline}
As shown in Figure \ref{fig:method}, the pipeline of FactSelfCheck consists of three main steps: (1) \textbf{Knowledge Graph Extraction} that extracts sets of entities, relations, and finally, knowledge graph from the initial response $p$ and samples $S$; (2) \textbf{Fact-level Hallucination Scoring}, that score facts by measuring factual consistency between facts in $KG_p$ and, depending on the variant, $KG_s$ in FactSelfCheck-KG or directly $s$ in FactSelfCheck-Text; (3) \textbf{Sentence-Level Score} calculation by aggregation of the fact-level scores.

\subsection{Knowledge Graph Extraction}

We adopt an approach that decomposes the knowledge graph extraction task into simpler subtasks, similarly to \citet{edge2024localglobalgraphrag}. This process involves three primary steps: extracting entities, identifying relations, and formulating facts, which are implemented as a sequence of LLM prompts.

For each instance, we extract a list of entities $\mathcal{E}_p$ by passing $p$ to the LLM.

\begin{equation}
    \mathcal{E}_p = LLM_{\text{entities}}(p)
\end{equation}

\noindent Next, we provide $p$ and $\mathcal{E}_p$ to the LLM to extract relation types between entities, resulting in $\mathcal{R}_p$.

\begin{equation}
    \mathcal{R}_p = LLM_{\text{relations}}(p, \mathcal{E}_p)
\end{equation}

\noindent We then input $\mathcal{E}_p$, $\mathcal{R}_p$, and each sentence $u \in U$ into the LLM to extract the knowledge graph $KG_u$. We also provide an initial response $p$ to add contextual information. The output is a set of facts:

\begin{equation}
    KG_u = LLM_{\text{sentence\_facts}}(u, p, \mathcal{E}_p, \mathcal{R}_p)
\end{equation}

\noindent After extracting $KG_u$ for each sentence $u \in U$, we compile the sets of entities $\mathcal{E}_S$ and relations $\mathcal{R}_S$ required for extracting knowledge graph $KG_s$ from each sample $s$ \footnote{Although we could theoretically use $\mathcal{E}_p$ and $\mathcal{R}_p$ for $KG_s$ extraction, in practice, LLMs are not sufficiently accurate to extract all entities and relations from the response $p$ when calculating $\mathcal{E}_p$ and $\mathcal{R}_p$. This results in $KG_u$ containing entities and relations not present in $\mathcal{E}_p$ and $\mathcal{R}_p$, even if the prompt restricts them. Empirical tests showed that extending $\mathcal{E}_S$ and $\mathcal{R}_S$ by adding entities and relations from all $KG_u$ improved the results.}:

\begin{equation}
    \mathcal{E}_S = \mathcal{E}_p \cup \bigcup_{u \in U} \{h, t \mid (h, r, t) \in KG_u\}
\end{equation}

\begin{equation}
    \mathcal{R}_S = \mathcal{R}_p \cup \bigcup_{u \in U} \{r \mid (h, r, t) \in KG_u\}
\end{equation}

\begin{equation}
    KG_s = LLM_{\text{sample\_facts}}(s, \mathcal{E}_S, \mathcal{R}_S)
\end{equation}

Extracting $KG_s$ by utilizing $\mathcal{E}_S$ and $\mathcal{R}_S$ is more convenient and robust than extraction without them, as it eliminates the need for entity alignment and ensures that the KG is built using the same schema as $KG_p$.

\subsection{Fact-Level Hallucination Scores}

We define two variants for measuring fact-level hallucination scores. The first variant, FactSelfCheck-KG, assesses the consistency between a fact and the knowledge graphs extracted from samples. The second variant, FactSelfCheck-Text, evaluates the consistency between a fact and the samples directly.

\subsubsection{FactSelfCheck-KG}

In the FactSelfCheck-KG variant, we introduce two metrics to assess the reliability of each fact.

\paragraph{Frequency-Based Hallucination Score}

The frequency-based fact-level hallucination score is based on the intuition that the probability of a fact being hallucinated is inversely proportional to the fraction of samples containing the same fact.

\begin{equation}
    \mathcal{H}_{\text{fact}}(f) = 1 - \frac{1}{|S|} \sum_{s \in S} \mathbb{I} \{ f \in KG_{s} \}
\end{equation}

\noindent where $\mathcal{H}_{\text{fact}}(f)$ is the hallucination score for fact $f$, and $\mathbb{I} \{ f \in KG_{s_n} \}$ is an indicator function that equals $1$ if fact $f$ appears in $KG_{s_n}$ and $0$ otherwise. The higher the $\mathcal{H}_{\text{fact}}$ value, the higher the plausibility of hallucination. 

\paragraph{LLM-Based Hallucination Score} 

To allow semantic matching and reasoning over knowledge graphs, rather than only exact fact matching, we introduce the LLM-based fact-level hallucination score. We instruct the LLM to determine whether each fact is supported by the knowledge graphs extracted from the samples. The LLM is expected to respond with 'yes' or 'no'. We then average the valid responses from the LLM to get the final score as in Equation~\eqref{Eq:factkghfact}. Any invalid responses are not included in the averaging.

\begin{equation}
    \label{Eq:factkghfact}
    \mathcal{H}_{\text{fact}}(f) = \frac{1}{|V_f|} \sum_{s \in V_f} \Psi(f, KG_{s})
\end{equation}

\noindent where $V_f$ represents the set of samples with valid LLM responses for the fact $f$, and the function $\Psi$ is defined as follows:

\begin{equation}
    \label{eq:psi}
    \Psi(\cdot) = \begin{cases}
        0 & \text{if the LLM returns 'yes'} \\
        1 & \text{if the LLM returns 'no'}
    \end{cases}
\end{equation}

\subsubsection{FactSelfCheck-Text}

In the FactSelfCheck-Text variant, we check if a fact is supported by each textual sample directly without using the knowledge graphs. We prompt the LLM to evaluate whether a fact $f$ is supported by the textual sample $s$. As in the previous variant, we average the valid LLM responses using the $\Psi$ function:

\begin{equation}
    \label{Eq:facttexthfact}
    \mathcal{H}_{\text{fact}}(f) = \frac{1}{|V_f|} \sum_{s \in V_f} \Psi(f, s)
\end{equation}

\subsection{Sentence-Level Hallucination Score}

\label{sec:method_sent_pass_level}

While detecting hallucinations at the fact level offers fine-grained insights, there are scenarios where sentence-level detection is necessary, such as for comparison with existing baselines. To achieve this, we aggregate fact-level scores to compute sentence-level scores $\mathcal{H}_{\text{sentence}}(u)$. This aggregation bridges the gap between atomic fact-level judgments and coarser sentence-level evaluations.

\begin{equation}
    \mathcal{H}_{\text{sentence}}(u) = \text{Agg}_{f \in KG_u} \mathcal{H}_{\text{fact}}(f)
\end{equation}

\noindent where $u$ represents a single sentence, and $U$ is the set of sentences of the response $p$. The aggregation function $\text{Agg}$ defines how the factuality of individual facts determines the factuality of the sentence. We employ two distinct aggregation strategies: $mean$ and $max$. The $mean$ function computes the average hallucination score of all facts within a sentence, providing a smoothed measure of the overall factual density. This is useful for assessing general content quality where partial correctness matters. In contrast, the $max$ function identifies the most severe hallucination within the sentence, based on the intuition that even a single hallucinated fact is sufficient to identify the sentence as hallucinated.

\section{Experimental Setup}
\label{sec:experiments}

In this section, we describe the experimental setup, including the used data, implementation details, and aspects we investigated. Additionally, we conducted a fact-level evaluation to provide a more comprehensive analysis. The detailed description of it is provided in Appendix \ref{sec:appendix_fact_level_evaluation}.

\subsection{Evaluation Data}
\label{sec:experiments-datasets}

Since the method is designed for detecting hallucinations in long generated passages, finding appropriate datasets was challenging. We evaluated our method on two datasets. The first one is the \nobreak{WikiBio} GPT-3 Hallucination Dataset \citep{manakul-etal-2023-selfcheckgpt}~\footnote{\url{huggingface.co/datasets/potsawee/wiki_bio_gpt3_hallucination} (cc-by-sa-3.0 license)} (later referred to as \nobreak{WikiBio}). To the best of our knowledge, this was the only dataset specifically designed for evaluating sampling-based hallucination detection methods, representing a significant research gap. To address this, we developed \nobreak{FavaMultiSamples}, a novel dataset specifically designed for evaluating methods that analyze multiple samples, providing researchers with an additional benchmark for robust evaluation. We built it upon the FAVA dataset \citep{mishra2024finegrained} and the detailed description of creation is provided in Appendix \ref{sec:appendix_fava}. \nobreak{WikiBio} covers generated biographical passages, while FavaMultiSamples includes diverse knowledge-intensive queries across various domains. Both datasets were annotated by humans. The dataset statistics are provided in Appendix~\ref{sec:appendix_datasets_stats}, and statistics related to KG extraction are detailed in Appendix~\ref{sec:appendix_steps_evaluation}. Both datasets contain only test data, making it methodologically incorrect to tune parameters, including classification thresholds, on these datasets. 

\paragraph{Sentence-level} While the datasets focus on sentence-level evaluation, our approach provides more fine-grained insights through fact-level analysis. To ensure a meaningful comparison with SelfCheckGPT, we evaluated these levels using aggregation approaches (see Section \ref{sec:method_sent_pass_level}). For WikiBio, we followed the protocol established by \citet{manakul-etal-2023-selfcheckgpt}, merging the labels \textit{major-inaccurate} and \textit{minor-inaccurate} into a single \textit{hallucination} class.

\paragraph{Fact-level} To enable fact-level evaluation, we annotated facts from the extracted response knowledge graph ($KG_p$) using the LLM-as-judge approach \citep{10.5555/3666122.3668142_llmasjudge}. For each fact, we provided the external biography from Wikipedia as a reference. The LLM-as-judge annotated whether each fact is supported by the reference. We utilized GPT-4o for this task, to ensure high quality of the annotations. As a result, we obtained $5488$ binary annotated facts.

The reliability of this approach is high, and commonly used in the literature \citep{thakur2025judgingjudgesevaluatingalignment}. The LLM-as-judge approach has been specifically tested for annotating hallucinations in LLM output given a knowledge source by \citet{janiak2025illusionprogressreevaluatinghallucination} and has demonstrated high alignment with human annotations, outperforming other automated evaluation methods such as ROUGE \citep{lin-2004-rouge}.

\subsection{Baseline Models}

We compared our method against several key baselines. The RandomSentence baseline predicts a random class for each sentence with equal probability. The RandomFact baseline predicts random scores for each fact and aggregates them to obtain sentence-level scores. The probability-based baselines, proposed by \citet{manakul-etal-2023-selfcheckgpt}, use $-$log$p$ and $\mathcal{H}$ to measure likelihood and entropy of each token respectively and aggregate the scores using Mean or Max functions to obtain sentence-level scores. 

We include the two best-performing variants of SelfCheckGPT: Prompt and NLI. We implemented the Prompt variant using the same LLM employed in our method, namely Llama-3.1-70B-Instruct, whereas the original method utilized an unspecified release of GPT-3.5-turbo. For the NLI variant, we used the same model as was used in the original paper\footnote{\url{huggingface.co/potsawee/deberta-v3-large-mnli} (Apache license 2.0)}. Finally, the AttentionScore leverages attention maps \citep{sriramanan2024llmcheck}, which is, to the best of our knowledge, the only unsupervised internal state-based method. Its unsupervised nature was crucial since the available datasets do not provide training data. We adapted the AttentionScore for sentence-level detection by implementing two variants: (1) Absolute, which analyzes the complete attention map from the input start to the end of the given sentence, and (2) Relative, which analyzes on the attention map between the start and end of the given sentence.

The AttentionScore and probability-based baselines are white-box methods, requiring access to the model. Since we cannot access all models used to generate the datasets, we passed the sequence of tokens to the proxy LLM to obtain the scores. The proxy LLM was the same as the one used in our method.

As discussed in Section \ref{sec:related}, fact-level methods like GraphEval and FactAlign require external knowledge source as reference, while GCA tuned hyperparameters on test data. These design differences make quantitative comparison with our zero-resource approach methodologically inappropriate.

\subsection{Implementation Details}
\label{sec:implementation_details}

We employed the Llama-3.1-70B-Instruct model \citep{grattafiori2024llama3herdmodels} as the LLM in all steps of our method and for baselines requiring access to LLM. We hosted it locally using vLLM \citep{kwon2023efficient} on a server with 2 x Nvidia H100 94GB. For annotation of facts and hallucination correction, we utilized GPT-4o \citep{openai2024gpt4ocard}, as motivated in Sections \ref{sec:experiments-datasets} and \ref{sec:exp-hallucination_correction}. We set the LLM's temperature to $0.0$ for all calls, except during hallucination correction (see Section \ref{sec:exp-hallucination_correction}), where we set it to $0.5$. The used prompts are available in the code repository. We implemented the methods and experiments using LangChain \citep{LangChain}, Hugging Face Transformers \citep{wolf-etal-2020-transformers} and Hugging Face Datasets \citep{lhoest-etal-2021-datasets}. All pipeline steps were defined using DVC \citep{ruslan_kuprieiev_2025_14636677_dvc} to facilitate reproducibility.

To determine whether a language model's response was 'yes' or 'no' in Equation \ref{eq:psi}, we parsed the text into individual words and verified the presence of the words 'yes' or 'no'. If either word was detected, the response was excluded from the averaging process in Equations \ref{Eq:factkghfact} and \ref{Eq:facttexthfact}. Moreover, our pipeline is vulnerable to not detecting facts in short, uninformative sentences; for these cases, we set the score to $0.5$. The analysis of the number of such sentences is provided in Appendix \ref{sec:appendix_steps_evaluation}.

\subsection{Sentence-Level Detection}

For a fair comparison, we employed the same evaluation protocol as SelfCheckGPT. We reported area under the precision-recall curve (AUC-PR). We ensured consistency in the evaluation protocol by reviewing their source code~\footnote{\url{github.com/potsawee/selfcheckgpt}}.

\subsection{Fact-Level Detection}

We evaluated all variants of FactSelfCheck using the fact-level score $\mathcal{H}_{\text{fact}}$. For comparison with SelfCheckGPT we used the best performing variant -- Prompt. As SelfCheckGPT provides only sentence-level granularity, we derived fact-level scores by averaging the sentence-level scores across all sentences containing each fact. 

\subsection{Role of Fact-Level Detection in Hallucination Correction}
\label{sec:exp-hallucination_correction}

One potential application of hallucination detection methods is their use in correcting hallucinated responses. In this experiment, we investigate the effectiveness of our fact-level detection approach in enhancing hallucination correction and compare the results with those obtained using sentence-level detection and a baseline method. Each of the three tested approaches uses different input for the LLM: \textbf{(1) Baseline}: the original prompt and the generated response. \textbf{(2) Sentence-level}: the original prompt, the generated response, and a list of hallucinated sentences. \textbf{(3) Fact-level}: the original prompt, the generated response, and a list of hallucinated facts.

As only the WikiBio dataset provides reference in form of the real Wikipedia biography, and the FavaMultiSamples dataset does not, we conducted this experiment only on WikiBio. The original prompt is the one used during the creation of the dataset: {\itshape "This is a Wikipedia passage about \{concept\_name\}:"}. We instructed the LLM to return a list of sentences, allowing it to correct each sentence or leave it unchanged if no hallucinations were detected. We obtained the lists of incorrect sentences/facts using the best variants of models on this dataset (see Section \ref{sec:results-sentence_doc}) with thresholds that achieved the highest F1-scores on the dataset ($0.3$ for FactSelfCheck-Text and $0.75$ for SelfCheckGPT (Prompt)).

Subsequently, we evaluated the factuality of the corrected responses using the LLM-as-judge approach \citep{10.5555/3666122.3668142_llmasjudge}. For each corrected sentence, we provided the external biography from Wikipedia as a reference. As elaborated in Section \ref{sec:experiments-datasets}, this approach has been tested for annotating hallucinations and demonstrates high alignment with human annotations \citep{janiak2025illusionprogressreevaluatinghallucination}. We instructed LLM-as-judge to return 'yes' if the source supported the sentence, 'no' if it was not, or 'refused' if the LLM declined to correct the sentence  (e.g., due to insufficient knowledge). We then categorized the responses into three labels: 'factual', 'non-factual', 'refused'. 

As mentioned in Section \ref{sec:implementation_details}, we utilized GPT-4o for correction and judging instead of \mbox{Llama-3.1-70B-Instruct} (used in detection). This choice was motivated by the challenging nature of the correction task -- the model needs to correct hallucinations using only its internal knowledge, without access to external references. While the model knows hallucinated parts, it must rely on its knowledge to determine the correct information.

While the method of correction described here is not our main contribution, we used it to study the potential benefits of fact-level detection. Although the correction method employed here may not be the most sophisticated, the key takeaway is the observed difference in performance.

\section{Results}
\label{sec:results}

This section presents the results of the experiments described in Section \ref{sec:experiments}. The additional results are presented in Appendix \ref{sec:appendix_additional}.

\subsection{Sentence-Level Detection}

\label{sec:results-sentence_doc}

\begin{table}[!htb]
    \centering
    \begin{tabularx}{\columnwidth}{ll@{\extracolsep{\fill}}r}
        \toprule
        \textbf{Method} & \textbf{Agg.} & \textbf{AUC-PR} \\
        \midrule
        \multicolumn{3}{c}{\textbf{Sentence-level methods}} \\
        SCGPT (Prompt) & - & 93.60 \\
        SCGPT (NLI) & - & 92.50 \\
        AttentionScore (Relative) & - & 83.85 \\
        Max($\mathcal{H}$) & - & 82.56 \\
        Mean($-$log$p$) & - & 79.20 \\
        Mean($\mathcal{H}$) & - & 79.02 \\
        Max($-$log$p$) & - & 78.41 \\
        AttentionScore (Absolute) & - & 77.95 \\
        RandomSentence & - & 72.96 \\
        \midrule
        \multicolumn{3}{c}{\textbf{Fact-level methods (ours)}} \\
        FSC-Text & max & 92.45 \\
        FSC-KG (LLM-based) & max & 91.82 \\
        FSC-Text & mean & 91.01 \\
        FSC-KG (LLM-based) & mean & 90.24 \\
        FSC-KG (Frequency-based) & max & 88.48 \\
        FSC-KG (Frequency-based) & mean & 88.25 \\
        RandomFact & mean & 74.22 \\
        \bottomrule
    \end{tabularx}
    \caption{WikiBio: Results on the sentence-level hallucination detection task. Comparison of sentence-level and fact-level methods based on AUC-PR scores. \uline{SCGPT} stands for SelfCheckGPT, \uline{FSC} represents FactSelfCheck, and \uline{Agg} denotes the aggregation method used for calculating sentence-level scores.}
    \label{tab:results_sentence}
\end{table}

\begin{table}[!htb]
    \centering
    \begin{tabularx}{\columnwidth}{ll@{\extracolsep{\fill}}r}
        \toprule
        \textbf{Method} & \textbf{Agg.} & \textbf{AUC-PR} \\
        \midrule
        \multicolumn{3}{c}{\textbf{Sentence-level methods}} \\
        SCGPT (Prompt) & - & 46.91 \\
        SCGPT (NLI) & - & 32.58 \\
        Max($\mathcal{H}$) & - & 28.22 \\
        Max($-$log$p$) & - & 26.20 \\
        AttentionScore (Relative) & - & 24.17 \\
        Mean($\mathcal{H}$) & - & 23.80 \\
        Mean($-$log$p$) & - & 22.85 \\
        AttentionScore (Absolute) & - & 22.19 \\
        RandomSentence & - & 21.70 \\
        \midrule
        \multicolumn{3}{c}{\textbf{Fact-level methods (ours)}} \\
        FSC-KG (Frequency-based) & max & 48.52 \\
        FSC-Text & max & 42.80 \\
        FSC-KG (LLM-based) & max & 40.63 \\
        FSC-Text & mean & 37.13 \\
        FSC-KG (Frequency-based) & mean & 36.16 \\
        FSC-KG (LLM-based) & mean & 35.81 \\
        RandomFact & mean & 21.22 \\
        \bottomrule
        \end{tabularx}
    \caption{FavaMultiSamples: Results on the sentence-level hallucination detection task. Comparison of sentence-level and fact-level methods based on AUC-PR scores. \uline{SCGPT} stands for SelfCheckGPT, \uline{FSC} represents FactSelfCheck, and \uline{Agg} denotes the aggregation method used for calculating sentence-level scores.}
    \label{tab:results_sentence_fava}
    \end{table}

Tables \ref{tab:results_sentence} and \ref{tab:results_sentence_fava} present a comparative analysis of our method against baselines.  For WikiBio, FactSelfCheck-Text utilizing $max$ as an aggregation function achieves an AUC-PR score of $92.45$. It demonstrates that our approach is comparable in performance to the leading SelfCheckGPT (SCGPT) variants -- Prompt ($93.60$) and NLI ($92.50$). Notably, while our method operates at a more granular level, it maintains competitive performance with a marginal decrease of $1.2\%$ compared to the best SCGPT. It is important to note that comparing our method to SelfCheckGPT at the sentence level inherently disadvantages our approach, as we operate at a more granular level of analysis. Nevertheless, our method still achieves competitive performance despite this inherent challenge.

For FavaMultiSamples, FactSelfCheck-KG (Frequency-based) with $max$ aggregation achieves the highest AUC-PR score of $48.52$, outperforming all sentence-level baselines, including SelfCheckGPT (Prompt) at $46.91$. This result, together with the findings from WikiBio, highlights that the best-performing FactSelfCheck (FSC) variant depends on the dataset. On WikiBio, FSC-Text, which performs direct comparison between facts and samples, consistently achieves the highest hallucination AUC-PR ($92.45$), outperforming FSC-KG, which relies on knowledge graph comparisons. FSC-Text offers computational advantages by eliminating the knowledge graph extraction step. However, on FavaMultiSamples, the frequency-based FSC-KG variant surpasses both FSC-Text and LLM-based FSC-KG. These differences suggest that the optimal variant is influenced by dataset characteristics, such as fact density, sentence length, and text style (e.g., the prevalence of lists in FavaMultiSamples). FSC-KG may be more computationally efficient for longer samples with lower fact density due to reduced token usage, while FSC-Text is preferable for shorter or denser samples. Regarding aggregation functions, $max$ outperformed $mean$ across both datasets. This makes intuitive sense -- a sentence is hallucinated if any fact within it is hallucinated.

An interesting side observation is that our reproduced SCGPT (Prompt) with Llama-3.1-70B-Instruct marginally surpassed the original implementation using GPT-3.5-turbo ($93.60$ vs $93.42$~\footnote{\label{footnote:original-scores}The scores on GPT-3.5-turbo were obtained from the original paper \citep{manakul-etal-2023-selfcheckgpt}.}).

\subsection{Fact-Level Detection}

\begin{table}[!htb]
    \centering

    \begin{tabularx}{\columnwidth}{l@{\extracolsep{\fill}}r}
        \toprule
        \textbf{Method} & \textbf{AUC-PR} \\
        \midrule
        FactSelfCheck-Text & 93.41 \\
        FactSelfCheck-KG (LLM-based) & 92.25 \\
        FactSelfCheck-KG (Freq.-based) & 87.99 \\
        SelfCheckGPT (Prompt) & 86.18 \\
        RandomFact & 65.79 \\
        \bottomrule
        \end{tabularx}
    \caption{Results on the fact-level hallucination detection task. Comparison of sentence-level and fact-level methods based on AUC-PR scores.}
    \label{tab:results_fact}
    \end{table}

Table \ref{tab:results_fact} presents the comparative results for \mbox{fact-level} hallucination detection on WikiBio. FactSelfCheck-Text demonstrates superior performance with an AUC-PR score of $93.41$, followed by FactSelfCheck-KG (LLM-based) at $92.25$. The frequency-based method achieves $87.99$, while SelfCheckGPT (Prompt) scores $86.18$. These results demonstrate the effectiveness of our method in detecting hallucinations at the fact level. Furthermore, the lower performance of averaging the sentence-level scores highlights the importance of designing fact-level methods and validating our approach.

Comparing FactSelfCheck to SelfCheckGPT could be seen as unfair because the latter operates at a lower granularity than required by the evaluation task. However, this situation is analogous to the sentence-level evaluation presented in Section \ref{sec:results-sentence_doc}. In both cases, direct comparisons are not fully appropriate. The key difference is that, at the fact level, FactSelfCheck significantly outperforms SelfCheckGPT, while at the sentence level, FactSelfCheck remains competitive despite the unfavorable comparison.

\subsection{Role of Fact-Level Detection in Hallucination Correction}
\label{sec:results-hallucination_correction}

Table \ref{tab:results_correction} presents the results of our hallucination correction experiment. The fact-level approach shows substantial improvements over the baseline and sentence-level methods. We observed a $35.5\%$ increase in factual content and $12.5\%$ reduction in non-factual content compared to the baseline. In contrast, the sentence-level detection achieves only modest improvements of $10.6\%$ and $4.8\%$, respectively, indicating that pointing out hallucinations at the fact level enables more effective corrections and underscoring the importance of our study and contributions. 

The overall rate of refusals remains low, increasing only marginally from $0.04$ in the baseline to $0.05$ with fact-level and sentence-level detection. We hypothesize that the model becomes more cautious with provided information about hallucinations and more likely to know its limitations. 

\begin{table}[!htb]
    \resizebox{\columnwidth}{!}{
\centering
\begin{tabular}{llll}
    \toprule
    \textbf{Level} & \textbf{Factual $\uparrow$} & \textbf{Non-Factual $\downarrow$} & \textbf{Refused} \\
    \midrule
    baseline & 0.23 & 0.74 & 0.04 \\
    sentence & 0.25 (+10.6\%) & 0.70 (-4.8\%) & 0.05 (+30.0\%) \\
    fact & 0.31 (+35.5\%) & 0.64 (-12.5\%) & 0.05 (+30.0\%) \\
    \bottomrule
    \end{tabular}
    }
    \caption{Effectiveness of hallucination correction by providing detected hallucinations at sentence-level, fact-level, and a baseline (without providing any hallucinations). The table presents the proportions of factual, non-factual sentences, and refused corrections. Percentages in parentheses indicate the relative change compared to the baseline. Arrows $\uparrow$ and $\downarrow$ denote whether a higher or lower value is better.}
    \label{tab:results_correction}
\end{table}

\section{Conclusion}
\label{sec:conclusion}

We introduced FactSelfCheck, a fact-level hallucination detection approach that achieves competitive performance with existing methods while providing more interpretable insights through structured knowledge representation. By detecting hallucinations at the granular fact level rather than sentence level, our method enables more effective hallucination correction through precise identification of incorrect facts. The zero-resource nature of our approach makes it broadly applicable across diverse domains without requiring external knowledge bases or domain-specific training data. Additionally, we contributed FavaMultiSamples, a novel benchmark addressing the critical gap in evaluation datasets for sampling-based hallucination detection methods. 

\section*{Limitations}
\label{sec:limitations}

Our study faces three primary limitations. First, we are constrained by the availability of suitable datasets. Although we contributed FavaMultiSamples, the second dataset for evaluating sampling-based methods, it is still a dataset with annotations at the sentence-level. The lack of datasets with long generated passages and annotations at the fact-level forced us to evaluate our method through aggregation rather than directly assessing our fact-level detection capabilities. Second, while the granular approach of FactSelfCheck justifies its increased complexity, the multiple LLM-based steps make it more computationally intensive compared to more straightforward methods like SelfCheckGPT. Third, our method could face challenges with very short or uninformative sentences where fact extraction may fail.

Several promising directions could address these limitations. An important step would be creating new datasets with fact-level hallucination annotations, enabling direct evaluation of our method's core capabilities. Additionally, we see significant potential for improving computational efficiency. Our current prompt engineering was largely empirical and not optimized for token usage. Future work could focus on reducing prompt lengths and merging steps, such as merging the KG extraction steps or simultaneously assessing support for multiple facts. 

\section*{Ethical Considerations}
\label{sec:ethics}

Like all machine learning methods, FactSelfCheck can produce false positives and false negatives. Therefore, it should not completely replace human verification of factual correctness in LLM responses.

\section*{Acknowledgments}
\label{sec:acknowledgments}

This work was funded by the European Union under the Horizon Europe grant OMINO – Overcoming Multilevel INformation Overload (grant number 101086321, \url{http://ominoproject.eu/}). Views and opinions expressed are those of the authors alone and do not necessarily reflect those of the European Union or the European Research Executive Agency. Neither the European Union nor the European Research Executive Agency can be held responsible for them. It was also co-financed with funds from the Polish Ministry of Education and Science under the programme entitled International Co-Financed Projects, grant no. 573977.  This work was co-funded by the National Science Centre, Poland under CHIST-ERA Open \& Re-usable Research Data \& Software  (grant number 2022/04/Y/ST6/00183). The infrastructure was provided by the CLARIN-PL project financed as part of the investment: "CLARIN ERIC – European Research Infrastructure Consortium: Common Language Resources and Technology Infrastructure (period: 2024-2026) funded by the Polish Ministry of Science and Higher Education (Programme: "Support for the participation of Polish scientific teams in international research infrastructure projects"), agreement number 2024/WK/01.

\bibliography{main}

\clearpage

\appendix

\section*{Appendix}

This appendix provides supplementary material organized as follows:

\textbf{Appendix \ref{sec:appendix_fava}:} FavaMultiSamples dataset creation and annotation methodology.

\textbf{Appendix \ref{sec:appendix_datasets_stats}:} Comprehensive statistics for both evaluation datasets.

\textbf{Appendix \ref{sec:appendix_additional}:} Additional results including complete sentence-level metrics with precision-recall curves, sample size ablation study, and evaluation of the intermediate steps of the pipeline. As well as confusion matrices and prediction statistics for the fact-level evaluation.

\textbf{Appendix \ref{sec:appendix_complexity}:} Computational complexity analysis.

\textbf{Appendix \ref{sec:appendix_case_study}:} Concrete example with knowledge graph visualization demonstrating fact-level detection advantages.

\textbf{Appendix \ref{sec:appendix_enhanced_prompt}:} Enhanced prompt experiment validating comparison fairness with SelfCheckGPT.

\section{FavaMultiSamples Dataset}
\label{sec:appendix_fava}

Addressing the lack of evaluation benchmarks for sampling-based hallucination detection methods, we built FavaMultiSamples upon the FAVA dataset\footnote{\url{https://huggingface.co/datasets/fava-uw/fava-data} license: CC-BY-4.0} developed by \citet{mishra2024finegrained}. Prior to our contribution, WikiBio was the only available dataset for evaluating sampling-based methods, limiting evaluation of different approaches. The original FAVA dataset contains $460$ passages generated by GPT (\mbox{gpt-3.5-turbo-0301}) and \mbox{Llama2-Chat-70B} in response to diverse information-seeking prompts. Each passage was annotated by trained annotators for factual accuracy. For more details about the dataset construction and annotation process, please refer to the original paper.

To create FavaMultiSamples, we generated $20$ samples for each passage with the temperature of $1.0$, matching the sample settings used in the WikiBio dataset. We used the same models that produced the original responses, with one exception: since gpt-3.5-turbo-0301 is no longer available, we used gpt-3.5-turbo-1106, the most similar model available at the time. The FAVA dataset uses an HTML-like format for annotations, so we split each generated response into sentences and annotated them in binary format, where $1$ indicates a sentence containing a hallucination.

\section{Dataset Statistics}
\label{sec:appendix_datasets_stats}

The statistics of the used datasets are presented in Table \ref{tab:datasets_stats}.

\begin{table}[!htb]
    \centering
    \begin{tabularx}{\columnwidth}{l@{\extracolsep{\fill}}rr}
        \toprule
         & \multirow{2}{*}{\textbf{WikiBio}} & \textbf{FavaMulti-} \\
         &  & \textbf{Samples} \\
        \midrule
        \# Passages & 238 & 460 \\
        \# Sentences & 1908 & 5660 \\
        \# Hall. sentences & 1392 & 1228 \\
        \# Fact. sentences & 516 & 4432 \\
        \% Hall. sentences & 72.96 & 21.70 \\
        Avg. sent./passage & 8.02 & 12.30 \\
        Avg. tok./passage & 184.77 & 340.72 \\
        Avg. tok./sentence & 23.48 & 30.30 \\
        \bottomrule
    \end{tabularx}
    \caption{Statistics of the used datasets: WikiBio and FavaMultiSamples. We summarize the number of passages, sentences, tokens and annotation statistics. The number of tokens were calculated using appropriate tokenizers for each model.}
    \label{tab:datasets_stats}
\end{table}

\section{Additional Results}
\label{sec:appendix_additional}

\subsection{Sentence-Level Detection}

\begin{table*}[!htb]
    \centering
    \begin{tabularx}{0.75\textwidth}{llccc}
        \toprule
        \multirow{2}{*}{\textbf{Method}} & \multirow{2}{*}{\textbf{Agg.}} & \multicolumn{3}{c}{\textbf{AUC-PR}} \\
        \cmidrule(lr){3-5}
        & & \textbf{Hallucination} & \textbf{Factuality} & \textbf{Avg.} \\
        \midrule
        \multicolumn{5}{c}{\textbf{Sentence-level methods}} \\
        SCGPT (Prompt) & - & 93.60 & 74.30 & 83.95 \\
        SCGPT (NLI) & - & 92.50 & 66.08 & 79.29 \\
        AttentionScore (Relative) & - & 83.85 & 51.62 & 67.74 \\
        Max($\mathcal{H}$) & - & 82.56 & 41.80 & 62.18 \\
        Mean($-$log$p$) & - & 79.20 & 44.11 & 61.65 \\
        Mean($\mathcal{H}$) & - & 79.02 & 49.18 & 64.10 \\
        Max($-$log$p$) & - & 78.41 & 33.59 & 56.00 \\
        AttentionScore (Absolute) & - & 77.95 & 42.23 & 60.09 \\
        RandomSentence & - & 72.96 & 27.04 & 50.00 \\
        \midrule
        \multicolumn{5}{c}{\textbf{Fact-level methods (ours)}} \\
        FSC-Text & max & 92.45 & 65.55 & 79.00 \\
        FSC-KG (LLM-based) & max & 91.82 & 64.64 & 78.23 \\
        FSC-Text & mean & 91.01 & 63.77 & 77.39 \\
        FSC-KG (LLM-based) & mean & 90.24 & 62.95 & 76.60 \\
        FSC-KG (Frequency-based) & max & 88.48 & 53.86 & 71.17 \\
        FSC-KG (Frequency-based) & mean & 88.25 & 55.27 & 71.76 \\
        RandomFact & mean & 74.22 & 29.74 & 51.98 \\
        \bottomrule
    \end{tabularx}
    \caption{WikiBio: Extended results on the sentence-level hallucination detection task. Comparison of sentence-level and fact-level methods based on AUC-PR scores for Hallucination, Factuality, and their Average. \uline{SCGPT} stands for SelfCheckGPT, \uline{FSC} represents FactSelfCheck, and \uline{Agg} denotes the aggregation method used for calculating sentence-level scores. The results are sorted by AUC-PR scores for Hallucination.}
    \label{tab:results_sentence_extended}
\end{table*}

\begin{table*}[!htb]
    \centering
    \begin{tabularx}{0.75\textwidth}{llccc}
        \toprule
        \multirow{2}{*}{\textbf{Method}} & \multirow{2}{*}{\textbf{Agg.}} & \multicolumn{3}{c}{\textbf{AUC-PR}} \\
        \cmidrule(lr){3-5}
        & & \textbf{Hallucination} & \textbf{Factuality} & \textbf{Avg.} \\
        \midrule
        \multicolumn{5}{c}{\textbf{Sentence-level methods}} \\
        SCGPT (Prompt) & - & 46.91 & 88.39 & 67.65 \\
        SCGPT (NLI) & - & 32.58 & 85.64 & 59.11 \\
        Max($\mathcal{H}$) & - & 28.22 & 81.90 & 55.06 \\
        Max($-$log$p$) & - & 26.20 & 80.84 & 53.52 \\
        AttentionScore (Relative) & - & 24.17 & 82.12 & 53.15 \\
        Mean($\mathcal{H}$) & - & 23.80 & 80.62 & 52.21 \\
        Mean($-$log$p$) & - & 22.85 & 80.38 & 51.62 \\
        AttentionScore (Absolute) & - & 22.19 & 81.71 & 51.95 \\
        RandomSentence & - & 21.70 & 78.30 & 50.00 \\
        \midrule
        \multicolumn{5}{c}{\textbf{Fact-level methods (ours)}} \\
        FSC-KG (Frequency-based) & max & 48.52 & 80.21 & 64.36 \\
        FSC-Text & max & 42.80 & 86.15 & 64.47 \\
        FSC-KG (LLM-based) & max & 40.63 & 85.01 & 62.82 \\
        FSC-Text & mean & 37.13 & 86.09 & 61.61 \\
        FSC-KG (Frequency-based) & mean & 36.16 & 79.92 & 58.04 \\
        FSC-KG (LLM-based) & mean & 35.81 & 84.65 & 60.23 \\
        RandomFact & mean & 21.22 & 77.77 & 49.50 \\
        \bottomrule
    \end{tabularx}
    \caption{FavaMultiSamples: Extended results on the sentence-level hallucination detection task. Comparison of sentence-level and fact-level methods based on AUC-PR scores for Hallucination, Factuality, and their Average. \uline{SCGPT} stands for SelfCheckGPT, \uline{FSC} represents FactSelfCheck, and \uline{Agg} denotes the aggregation method used for calculating sentence-level scores. The results are sorted by AUC-PR scores for Hallucination.}
    \label{tab:results_sentence_fava_extended}
\end{table*}

Tables \ref{tab:results_sentence_extended} and \ref{tab:results_sentence_fava_extended} provide a more comprehensive view of the FactSelfCheck variants against baselines on both datasets, including AUC-PR scores for detecting factual sentences (Factuality AUC-PR) in addition to hallucinated ones, and an average of these two. While Section \ref{sec:results-sentence_doc} focused on hallucination detection, these extended results offer additional insights into our approach. On WikiBio, FSC-Text (max aggregation) shows comparable performance in hallucination detection with $92.45$ AUC-PR vs. $93.60$ for SCGPT (Prompt). While SCGPT (Prompt) achieves a higher factuality detection score ($74.30$ vs. $65.55$), our method performs well in identifying potential misinformation and contributes to effective factual verification.

On the FavaMultiSamples dataset, FSC-KG (Frequency-based, max aggregation) demonstrates high overall performance, achieving a factuality detection AUC-PR of $80.21$ alongside solid hallucination detection ($48.52$ AUC-PR compared to SCGPT Prompt's $46.91$). This performance across metrics shows the adaptability of our approach to different datasets. While SCGPT (Prompt) achieves a higher average AUC-PR ($67.65$ vs. $64.36$) due to stronger factuality detection ($88.39$), our method provides balanced performance across both metrics, offering advantages for applications where hallucination detection is important.

\begin{figure}[!htb]
    \centering
    \includegraphics[width=0.85\columnwidth]{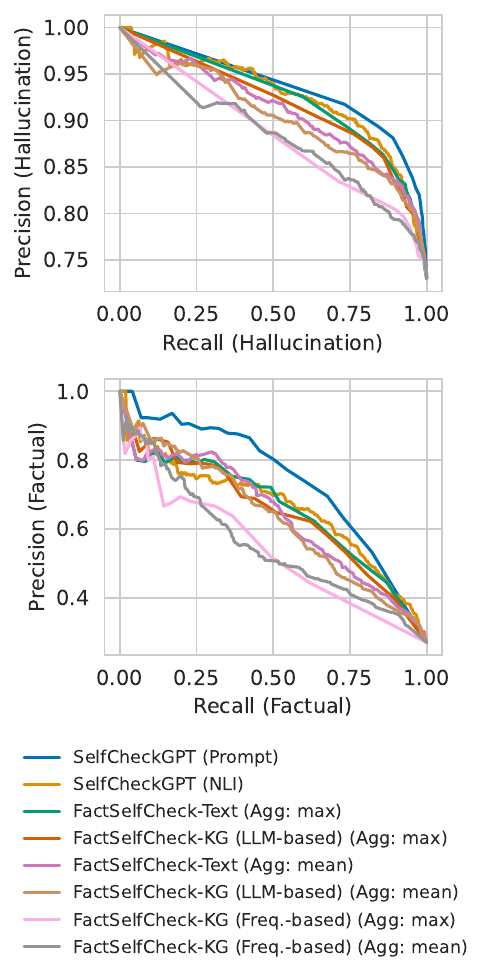}
    \caption{WikiBio: Precision-recall curve for the sentence-level hallucination and factuality detection.}
    \label{fig:pr_curve}
\end{figure}

\begin{figure}[!htb]
    \centering
    \includegraphics[width=0.85\columnwidth]{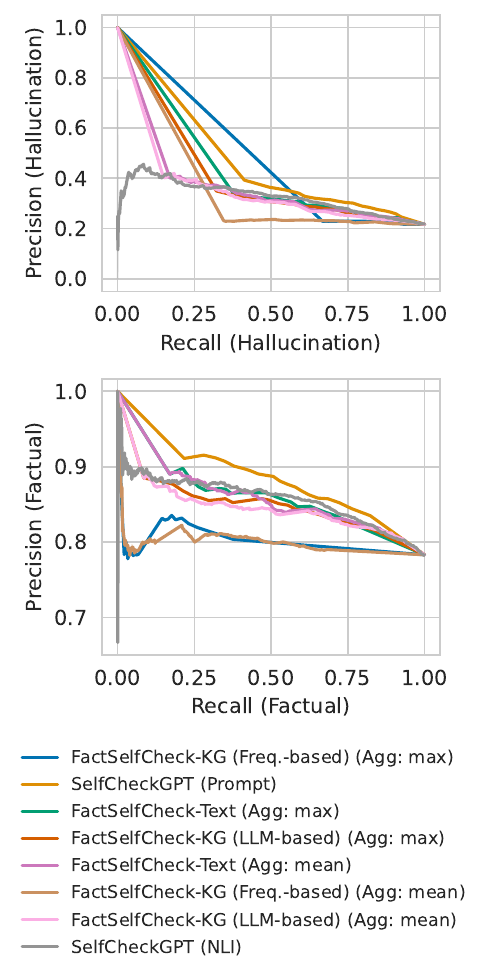}
    \caption{FavaMultiSamples: Precision-recall curve for the sentence-level hallucination and factuality detection.}
    \label{fig:pr_curve_fava}
\end{figure}

Figures \ref{fig:pr_curve} and \ref{fig:pr_curve_fava} illustrate the precision-recall curves for the sentence-level detection task on the WikiBio and FavaMultiSamples datasets, respectively. These curves show the performance across various thresholds for both hallucination and factuality detection.

\subsection{Fact-Level Detection}
\label{sec:appendix_fact_level_evaluation}

\begin{table}[!htb]
    \centering
    \begin{tabularx}{\columnwidth}{l@{\extracolsep{\fill}}rr}
        \toprule
        & \textbf{Pred Fact.} & \textbf{Pred Hall.} \\
       \midrule
       \textbf{True Fact.} & 951 & 832 \\
       \textbf{True Hall.} & 175 & 3530 \\
       \bottomrule
    \end{tabularx}
    \caption{Confusion matrix of FactSelfCheck-Text on fact-level evaluation.}
\label{tab:results_fact_fsc_cm}
\end{table}

\begin{table}[!htb]
    \centering
    \begin{tabularx}{\columnwidth}{l@{\extracolsep{\fill}}rr}
        \toprule
        & \textbf{Pred Fact.} & \textbf{Pred Hall.} \\
       \midrule
       \textbf{True Fact.} & 570 & 1213 \\
       \textbf{True Hall.} & 202 & 3503 \\
       \bottomrule
    \end{tabularx}
    \caption{Confusion matrix of SelfCheckGPT (Prompt) on fact-level evaluation.}
\label{tab:results_fact_scgpt_cm}
\end{table}

\begin{table}[!htb]
    \centering
    \begin{tabularx}{\columnwidth}{l@{\extracolsep{\fill}}rr}
        \toprule
        \textbf{Method} & \textbf{Correct} & \textbf{Incorrect} \\
        \midrule
        SelfCheckGPT & 4073 & 1415 \\
        FactSelfCheck & 4481 & 1007 \\
        \bottomrule
    \end{tabularx}
    \caption{Overall prediction accuracy comparison for fact-level evaluation.}
\label{tab:results_fact_incorrect_pred}
\end{table}

To extend the demonstration of the effectiveness of our fact-level detection approach, we present confusion matrices and prediction statistics. Table~\ref{tab:results_fact_fsc_cm} shows the confusion matrix for FactSelfCheck-Text, while Table~\ref{tab:results_fact_scgpt_cm} presents results for SelfCheckGPT (Prompt) when adapted to fact-level prediction. 

FactSelfCheck-Text correctly identifies $3530$ hallucinated facts and $951$ factual facts, achieving higher true positive rates for hallucination detection compared to SelfCheckGPT, which correctly identifies $3503$ hallucinated facts but only $570$ factual facts.

Table \ref{tab:results_fact_incorrect_pred} summarizes the overall prediction accuracy of both methods. FactSelfCheck achieves $4481$ correct predictions compared to $4073$ for SelfCheckGPT, representing an increase of $408$ correct predictions ($10.02\%$ improvement).

\subsection{Effect of Sample Size on Detection Performance}
\label{app:sample_size}

We investigated the impact of varying the number of samples on the detection performance of our method. Specifically, we evaluated the performance at the sentence level by changing the number of samples from $1$ to $20$. We compared our method with SelfCheckGPT.

\begin{figure}[!htb]
    \centering
    \includegraphics[width=0.85\columnwidth]{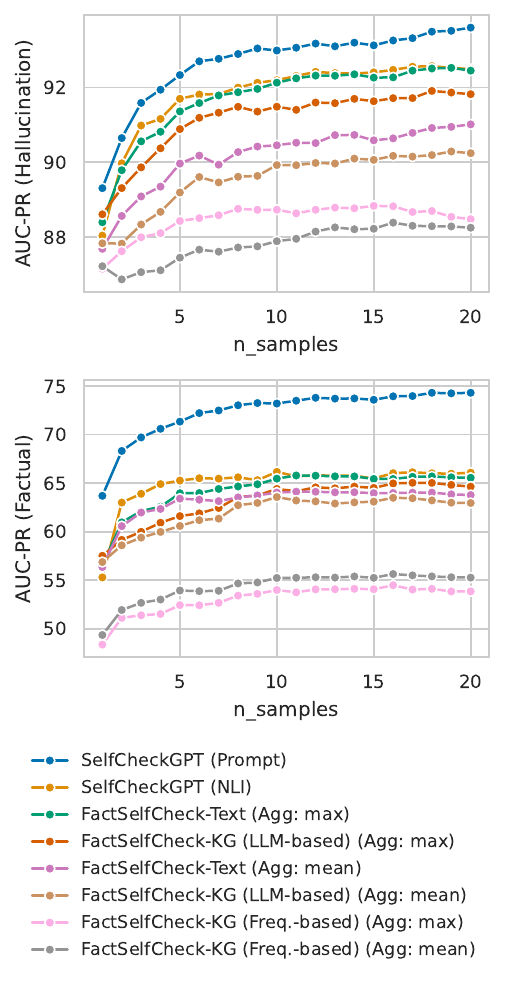}
    \caption{WikiBio: Impact of sample size on both hallucination and factuality detection performance for different methods.}
    \label{fig:ablation_n_samples_sentence_extended}
\end{figure}

\begin{figure}[!htb]
    \centering
    \includegraphics[width=0.85\columnwidth]{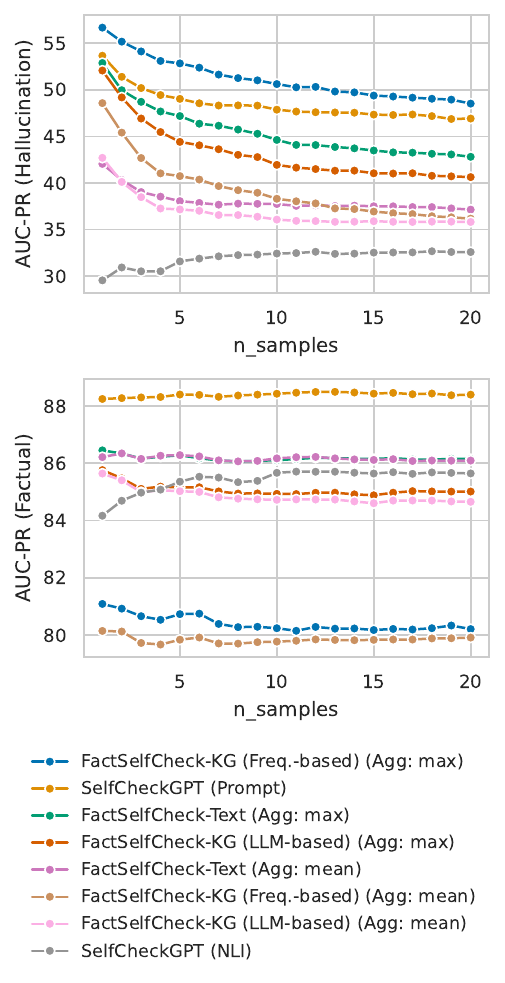}
    \caption{FavaMultiSamples: Impact of sample size on both hallucination and factuality detection performance for different methods.}
    \label{fig:ablation_n_samples_sentence_fava_extended}
\end{figure}

Figure \ref{fig:ablation_n_samples_sentence_extended} illustrates that that on WikiBio, FactSelfCheck exhibits similar behavior to SelfCheckGPT regarding sample requirements. The performance improves dramatically with up to $5$ samples, after which the improvement curve flattens. While additional samples continue to yield benefits, these improvements become incremental, with modest gains observed up to $20$ samples. Factuality detection exhibits similar patterns to hallucination detection. This pattern confirms that more samples provide better evidence for accurate detection across both metrics. 

Figure \ref{fig:ablation_n_samples_sentence_fava_extended} shows a different pattern for FavaMultiSamples, where hallucination detection performance decreases with more samples. This occurs because methods with few samples tend to overestimate hallucination scores, interpreting normal variations as potential hallucinations. As sample size increases, methods become more conservative in their scoring, leading to better calibration. This pattern likely stems from FavaMultiSamples having shorter sentences with fewer facts compared to WikiBio (see Section \ref{sec:appendix_steps_evaluation}).

The distinct patterns between datasets highlight that sampling effects are context-dependent. While more samples universally improve calibration quality, the impact on raw performance metrics depends on dataset characteristics and initial score distributions. It's important to note that AUC-PR, while useful, has limitations. A classifier that consistently returns the same score can achieve a high AUC-PR value, which may not reflect true discriminative ability. Therefore, the decrease in AUC-PR with more samples might actually indicate better calibration and more meaningful score distributions, rather than worse performance.

\subsection{Evaluation of Intermediate Steps}
\label{sec:appendix_steps_evaluation}

Our method consists of multiple steps that cannot be directly evaluated due to the lack of human annotations. While previous sections evaluated the complete pipeline, to strengthen our study we also analyzed and validated statistics from intermediate steps. We examined the number of entities and relations per passage, along with facts per sentence, calculating mean, minimum, maximum values, and the count and percentage of entries with zero elements. Sentences with no facts are particularly important as FactSelfCheck assigns them a default score of $0.5$ (see Section \ref{sec:implementation_details}).

\begin{table}[!htb]
    \centering
    \begin{tabularx}{\columnwidth}{l@{\extracolsep{\fill}}rrr}
        \toprule
        & \textbf{\# entities/} & \textbf{\# relations/} & \textbf{\# facts/} \\
        & \textbf{passage} & \textbf{passage} & \textbf{sentence} \\
        \midrule
        \multicolumn{4}{c}{\textbf{WikiBio}} \\
        \midrule
        mean & 24.85 & 21.18 & 3.24 \\
        min & 11 & 5 & 0 \\
        max & 282 & 163 & 85 \\
        \# 0 el. & 0 & 0 & 22 \\
        \% 0 el. & 0.00\% & 0.00\% & 1.15\% \\
        \midrule
        \multicolumn{4}{c}{\textbf{FavaMultiSamples}} \\
        \midrule
        mean & 32.08 & 50.34 & 2.97 \\
        min & 1 & 0 & 0 \\
        max & 312 & 237 & 102 \\
        \# 0 el. & 0 & 2 & 344 \\
        \% 0 el. & 0.00\% & 0.43\% & 6.08\% \\
        \bottomrule
    \end{tabularx}
    \caption{Statistics of intermediate steps in FactSelfCheck across both datasets. The table shows the distribution of entities per passage, relations per passage, and facts per sentence, including occurrences of entries with 0 elements.}
    \label{tab:steps_evaluation}
    \end{table}

Table \ref{tab:steps_evaluation} reveals that both datasets have high mean numbers of entities and relations per passage, indicating that knowledge graph construction is not constrained by earlier steps. While WikiBio shows acceptable minimum values for entities and relations per passage, FavaMultiSamples exhibits notably lower minimums that could impact knowledge graph extraction performance. The percentage of sentences without extracted facts is relatively low in WikiBio ($1.15\%$) but more substantial in FavaMultiSamples ($6.08\%$), potentially affecting detection accuracy. Both datasets show considerable variability, with some passages containing over 200 entities and relations, highlighting the diverse complexity of the analyzed generated responses.

\section{Computational complexity}
\label{sec:appendix_complexity}

\begin{table}[!htb]
    \centering
    \begin{tabular}{ll}
        \toprule
        \textbf{Method} & \textbf{Number of LLM calls} \\
        \midrule
        FSC-KG (Freq.) & $2 + |U| + |S|$ \\
        FSC-KG (LLM) & $2 + |U| + |S| + |KG_p| \times |S|$ \\
        FSC-Text & $2 + |U| + |KG_p| \times |S|$ \\
        SCGPT-Prompt & $|U| \times |S|$ \\
        \bottomrule
    \end{tabular}
    \caption{Computational complexity of different methods in terms of number of LLM calls required. $U$ is the set of sentences in the generated passage, $S$ is the set of stochastic LLM response samples, and $KG_p$ is the knowledge graph containing all extracted facts.}
    \label{tab:computational_complexity}
\end{table}

While FactSelfCheck is more granular than SelfCheckGPT, this comes with increased computational costs. In Table \ref{tab:computational_complexity} we compare all variants of FactSelfCheck with SelfCheckGPT (Prompt) in terms of the number of LLM calls required. Every variant of FactSelfCheck requires additional calls for entity and relation extraction (constant $2$), followed by knowledge graph construction for each sentence ($|U|$), then it assesses factual consistency of each fact in different ways with varying complexity. As noted in Limitations, we did not optimize for token usage, and future work could merge steps to reduce complexity while maintaining the fine-grained insights our method provides.

\section{Case Study of FactSelfCheck vs SelfCheckGPT}
\label{sec:appendix_case_study}

\begin{figure*}[!htb]
    \centering
    \includegraphics[width=0.8\textwidth,fbox]{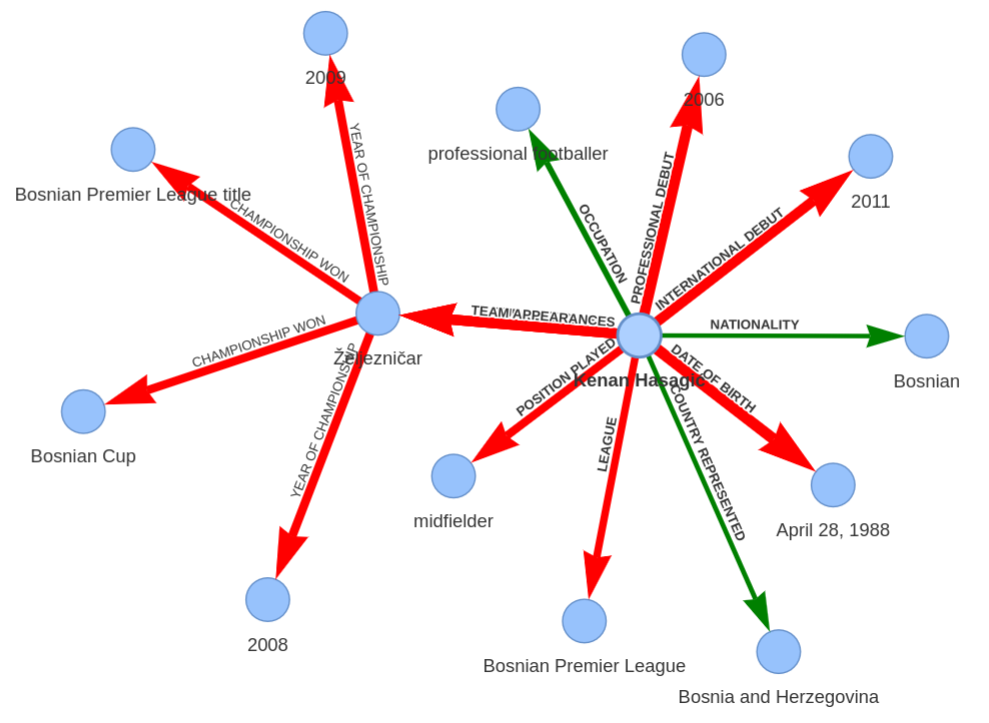}
    \caption{Example of a knowledge graph extracted from a response. Edge width represents the hallucination score for each fact, with red edges indicating hallucinated facts and green edges indicating correct facts. Facts were classified using a threshold of $0.3$, which achieved the highest F1-score in the fact-level evaluation (\ref{sec:appendix_fact_level_evaluation}).}
    \label{fig:example}
\end{figure*}

Table \ref{tab:example} and Figure \ref{fig:example} present a comparative case study of predictions from WikiBio made by FactSelfCheck and SelfCheckGPT. The table contains an external Wikipedia biography, sentences from the response, facts extracted from the response, and the predictions of both methods. This comparison demonstrates that fact-level detection provides more detailed information about the factuality of the response. We observe that LLM did not hallucinate all facts, as some are consistent with the external Wikipedia biography. However, when using sentence-level detection, we cannot distinguish between correct and hallucinated facts -- all sentences are predicted as hallucinated.

\begin{table*}[!htb]
    \small
    \centering
    \begin{tabularx}{0.98\textwidth}{l@{\extracolsep{\fill}}c}
    \toprule
    \multicolumn{2}{c}{\textbf{External Wikipedia Bio}} \\
    \multicolumn{2}{L{0.95\textwidth}}{Kenan Hasagić (born 1 February 1980) is a Bosnian football goalkeeper who plays for Balıkesirspor. His football career began in his hometown with FK Rudar. At the age of 16, he made his debut in a first division match. He was the most promising goalkeeper in Bosnia and Herzegovina; he played for youth selections and was later transferred to Austrian side Vorwärts Steyr. After that, he was a member of Altay SK in Turkey but didn't see much first team football. He went back to Bosnia and played for Bosna Visoko. In 2003, he signed a contract with FK Željezničar. Here he found good form and even became first choice goalkeeper for the Bosnian national team. In the 2004–05 season, he moved to Turkey once again where he signed for Turkish Süper Lig side Gaziantepspor. He made his debut for the national team on 12 February 2003 in a game between Wales and Bosnia and Herzegovina which ended in a 2–2 draw.} \\
    \midrule
    \multicolumn{2}{c}{\textbf{Sentence 1} (SelfCheckGPT: \textcolor{red}{1.0})} \\        
    \multicolumn{2}{L{0.95\textwidth}}{Kenan Hasagić (born 28 April 1988) is a Bosnian professional footballer who plays as a midfielder for Bosnian Premier League club Željezničar.} \\ 
    \multicolumn{1}{c}{\textbf{Fact}} & \textbf{FactSelfCheck} \\
    ('Kenan Hasagić', 'DATE OF BIRTH', 'April 28, 1988') & \textcolor{red}{1.00} \\
    ('Kenan Hasagić', 'NATIONALITY', 'Bosnian') & \textcolor{ForestGreen}{0.00} \\
    ('Kenan Hasagić', 'OCCUPATION', 'professional footballer') & \textcolor{ForestGreen}{0.20} \\
    ('Kenan Hasagić', 'POSITION PLAYED', 'midfielder') & \textcolor{red}{0.55} \\
    ('Kenan Hasagić', 'LEAGUE', 'Bosnian Premier League') & \textcolor{red}{0.45} \\
    ('Kenan Hasagić', 'CURRENT CLUB', 'Željezničar') & \textcolor{red}{0.85} \\
    \midrule
    \multicolumn{2}{c}{\textbf{Sentence 2} (SelfCheckGPT: \textcolor{red}{1.0})} \\
    \multicolumn{2}{L{0.95\textwidth}}{Hasagić started his career at his hometown club Željezničar, where he made his professional debut in 2006.} \\ 
    \multicolumn{1}{c}{\textbf{Fact}} & \textbf{FactSelfCheck} \\
    ('Kenan Hasagić', 'CURRENT CLUB', 'Željezničar') & \textcolor{red}{0.85} \\
    ('Kenan Hasagić', 'PROFESSIONAL DEBUT', '2006') & \textcolor{red}{0.90} \\
    ('Kenan Hasagić', 'CURRENT CLUB', 'Željezničar') & \textcolor{red}{0.85} \\
    \midrule
    \multicolumn{2}{c}{\textbf{Sentence 3} (SelfCheckGPT: \textcolor{red}{1.0})} \\
    \multicolumn{2}{L{0.95\textwidth}}{He has since gone on to make over 200 appearances for the club, winning the Bosnian Premier League title in 2008 and the Bosnian Cup in 2009.} \\ 
    \multicolumn{1}{c}{\textbf{Fact}} & \textbf{FactSelfCheck} \\
    ('Kenan Hasagić', 'TEAM APPEARANCES', 'Željezničar') & \textcolor{red}{0.60} \\
    ('Željezničar', 'CHAMPIONSHIP WON', 'Bosnian Premier League title') & \textcolor{red}{0.90}  \\
    ('Željezničar', 'YEAR OF CHAMPIONSHIP', '2008') & \textcolor{red}{1.00} \\
    ('Željezničar', 'CHAMPIONSHIP WON', 'Bosnian Cup') & \textcolor{red}{0.95} \\
    ('Željezničar', 'YEAR OF CHAMPIONSHIP', '2009') & \textcolor{red}{1.00} \\
    \midrule
    \multicolumn{2}{c}{\textbf{Sentence 4} (SelfCheckGPT: \textcolor{red}{0.9})} \\
    \multicolumn{2}{L{0.95\textwidth}}{He has also represented Bosnia and Herzegovina at international level, making his debut in 2011.} \\ 
    \multicolumn{1}{c}{\textbf{Fact}} & \textbf{FactSelfCheck} \\
    ('Kenan Hasagić', 'COUNTRY REPRESENTED', 'Bosnia and Herzegovina') & \textcolor{ForestGreen}{0.05} \\
    ('Kenan Hasagić', 'INTERNATIONAL DEBUT', '2011') & \textcolor{red}{0.85} \\
    \bottomrule
    \end{tabularx}
    \caption{Comparison of fact-level FactSelfCheck with sentence-level SelfCheckGPT. An external Wikipedia biography is provided to analyse the correctness of the methods. The red value indicates hallucinations, and the green value indicates factual correctness. The facts were classified using a threshold of $0.3$ utilizing FactSelfCheck, and the sentences were classified using a threshold of $0.75$ with SelfCheckGPT. These thresholds achieved the highest F1-scores in fact-level (Appendix \ref{sec:appendix_fact_level_evaluation}) and sentence-level (Section \ref{sec:results-sentence_doc}) evaluation, respectively.}
    \label{tab:example}
\end{table*}

\section{SelfCheckGPT with Enhanced Prompt}
\label{sec:appendix_enhanced_prompt}

To ensure a fair comparison between \nobreak{FactSelfCheck} and SelfCheckGPT, during preliminary studies, we conducted an additional experiment using an enhanced prompt for SelfCheckGPT. The original SelfCheckGPT prompt is relatively simple, while our FactSelfCheck prompts are more elaborate, directly allowing reasoning and inference of new facts, and providing examples. These characteristics could potentially increase the performance of methods. We designed an alternative prompt for SelfCheckGPT, that incorporates these features, making it similar to our FactSelfCheck prompts. 

For sentence-level detection on WikiBio, the enhanced prompt for SelfCheckGPT achieved an AUC-PR score of $93.38$, slightly lower than the original prompt's $93.60$. This minimal difference indicates that SelfCheckGPT's performance is not significantly affected by prompt design in our experimental setting. In fact, the enhanced prompt even lowered the performance rather than increasing it, despite its more sophisticated design.

Due to these findings, we chose to use the original prompt for SelfCheckGPT in our experiments. These consistent results confirm that our comparison between methods is fair, as the enhanced prompt did not improve the performance of SelfCheckGPT.

\end{document}